\pdfoutput=1

\documentclass[11pt]{article}

\usepackage[]{acl}

\usepackage{times}
\usepackage{latexsym}

\usepackage[T1]{fontenc}

\usepackage[utf8]{inputenc}

\usepackage{microtype}

\usepackage{amsmath}
\usepackage{graphicx}
\usepackage{CJKutf8}
\usepackage{multirow}
\usepackage{booktabs}
\usepackage{cleveref}
\usepackage{mwe}
\usepackage{amssymb}
\usepackage{subcaption}

\title{EASE: Entity-Aware Contrastive Learning of Sentence Embedding}

\author{
    Sosuke Nishikawa$^{\dagger,\ddagger}$\thanks{\hspace{2mm}Work done as an intern at Studio Ousia.}\\
    {\small \texttt{sosuke-nishikawa@alumni.u-tokyo.ac.jp}}
    \And
    Ryokan Ri$^{\dagger,\ddagger}$\\
    {\small \texttt{ryo0123@ousia.jp}}
    \AND
    Ikuya Yamada$^{\dagger,\mathsection}$\\
    {\small \texttt{ikuya@ousia.jp}}
    \And
    Yoshimasa Tsuruoka$^{\ddagger}$\\
    {\small \texttt{tsuruoka@logos.t.u-tokyo.ac.jp}}
    \And
    Isao Echizen$^{\ddagger,\mathparagraph}$\\
    {\small \texttt{iechizen@nii.ac.jp}}
    \AND
    \begin{minipage}{\textwidth}
        \begin{center}
            \fontsize{11.5}{14}\selectfont
            \textnormal{$^\dagger$Studio Ousia\,\,\,$^\ddagger$The University of Tokyo}\\
            \textnormal{$^\mathsection$RIKEN AIP\,\,\,$^\mathparagraph$National Institute of Informatics} \end{center}
    \end{minipage}
}

\begin{document}
\maketitle
\begin{abstract}
We present EASE, a novel method for learning sentence embeddings via contrastive learning between sentences and their related entities.
The advantage of using entity supervision is twofold: (1) entities have been shown to be a strong indicator of text semantics and thus should provide rich training signals for sentence embeddings; (2) entities are defined independently of languages and thus offer useful cross-lingual alignment supervision.
We evaluate EASE against other unsupervised models both in monolingual and multilingual settings.
We show that EASE exhibits competitive or better performance in English semantic textual similarity (STS) and short text clustering (STC) tasks and it significantly outperforms baseline methods in multilingual settings on a variety of tasks.
Our source code, pretrained models, and newly constructed multilingual STC dataset are available at \url{https://github.com/studio-ousia/ease}.

\end{abstract}

\newcommand{\Japanese}[1]{{\begin{CJK}{UTF8}{min}#1\end{CJK}}}

\newcommand{\Appendix}[1]{Appendix \ref{#1}}
\newcommand{\Table}[1]{Table \ref{#1}}
\newcommand{\Figure}[1]{Figure \ref{#1}}

\newcommand{\ie}{{\it i.e.}}
\newcommand{\eg}{{\it e.g.}}

\newcommand{\ba}{$\rm{_{base}}$}

\section{Introduction}
The current dominant approach to learning sentence embeddings is fine-tuning general-purpose pretrained language models, such as BERT \citep{devlin-etal-2019-bert}, with a particular training supervision.
The type of supervision can be natural language inference data \citep{reimers-gurevych-2019-sentence}, adjacent sentences \citep{yang-etal-2021-universal}, or a parallel corpus for multilingual models \citep{Feng2020LanguageagnosticBS}.

In this paper, we explore a type of supervision that has been under-explored in the literature: \emph{entity hyperlink annotations} from Wikipedia.
Their advantage is twofold: (1) entities have been shown to be a strong indicator of text semantics \citep{Gabrilovich2007ESA,yamada-etal-2017-learning,yamada-etal-2018-representation,Ling2020LearningCE} and thus should provide rich training signals for sentence embeddings;
(2) entities are defined independently of languages and thus offer a useful cross-lingual alignment supervision \citep{Calixto2021naacl,nishikawa2021multilingual,XLM-K-2021-AAAI,ri2021mluke}. The extensive multilingual support of Wikipedia alleviates the need for a parallel resource to train well-aligned multilingual sentence embeddings, especially for low-resource languages.
To demonstrate the effectiveness of entity-based supervision, we present {\bf EASE} (Entity-Aware contrastive learning of Sentence Embeddings), which produces high-quality sentence embeddings in both monolingual and multilingual settings.

\begin{figure}[t]
  \centering
  \includegraphics[width=7.8cm]{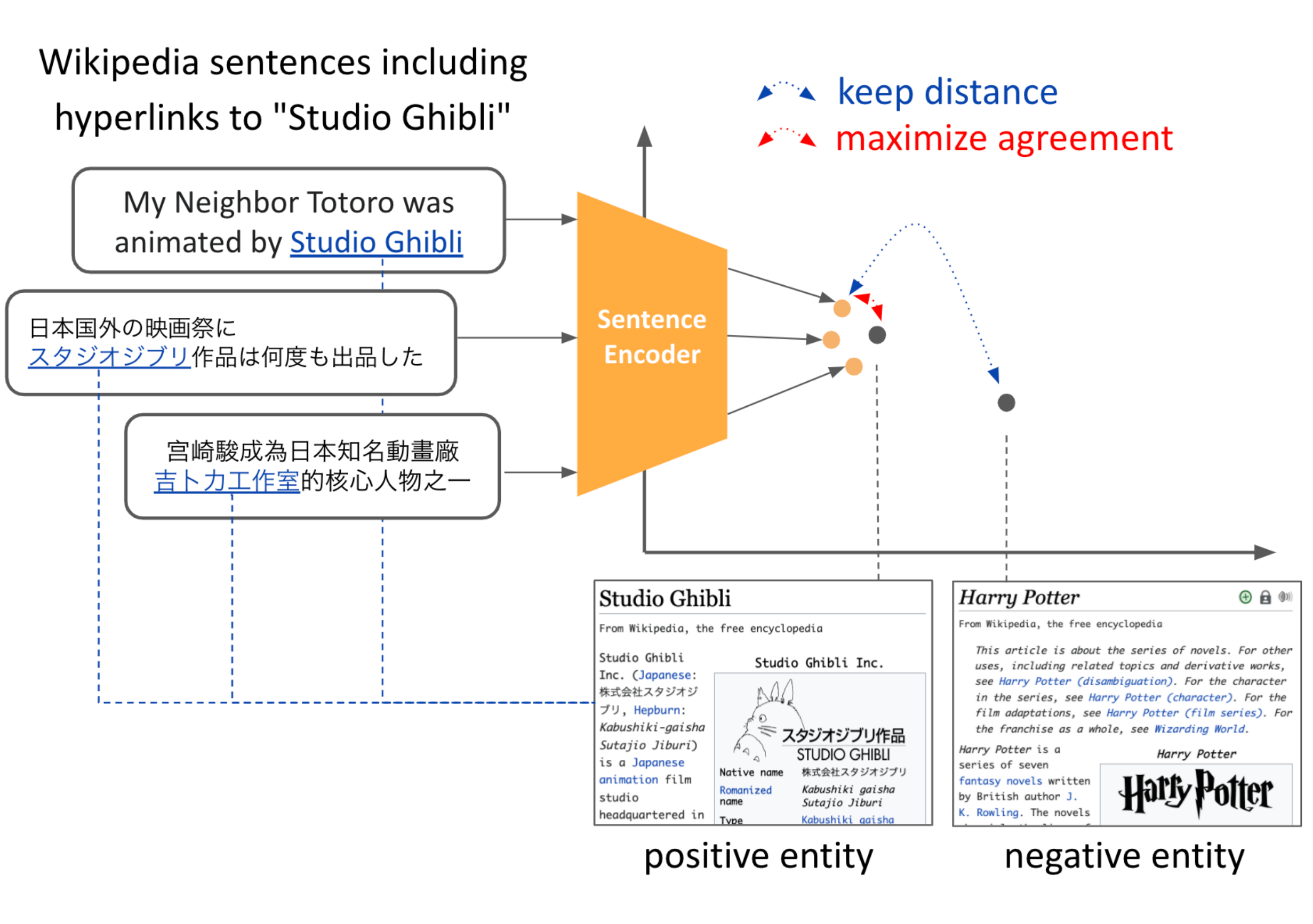}
  \caption{Illustration of the main concept behind EASE. Using a contrastive framework, sentences are embedded in the neighborhood of their hyperlink entity embeddings and kept apart from irrelevant entities. Here, we share the entity embeddings across languages for multilingual models to facilitate cross-lingual alignment of the representation.}
  \label{fg: ease}
\end{figure}

EASE learns sentence embeddings with two types of objectives: (1) our novel entity contrastive learning (CL) loss between sentences and their related entities (Figure \ref{fg: ease}); (2) the self-supervised CL loss with dropout noise.
The entity CL objective pulls the embeddings of sentences and their related entities close while keeping unrelated entities apart. The objective is expected to arrange the sentence embeddings in accordance with semantics captured by the entities.
To further exploit the knowledge in Wikipedia and improve the learned embeddings, we also introduce a method for mining hard negatives based on the entity type.
The second objective, the self-supervised CL objective with dropout noise \citep{gao-etal-2021-simcse,liu-etal-2021-fast}, is combined with the first one to enable sentence embeddings to capture fine-grained text semantics.
We evaluate our model against other state-of-the-art unsupervised sentence embedding models, and show that EASE exhibits competitive or better performance on semantic textual similarity (STS) and short text clustering (STC) tasks.

We also apply EASE to multilingual settings.
To facilitate the evaluation of the high-level semantics of multilingual sentence embeddings, we construct a multilingual text clustering dataset, MewsC-16 (Multilingual Short Text Clustering Dataset for News in 16 languages).
Multilingual EASE is trained using the entity embeddings shared across languages.
We show that, given the cross-lingual alignment supervision from the shared entities, multilingual EASE significantly outperforms the baselines in multilingual STS, STC, parallel sentence matching, and cross-lingual document classification tasks.

We further demonstrate the effectiveness of the multilingual entity CL in a more realistic scenario for low-resource languages.
Using multilingual entity CL, we fine-tune a competitive multilingual sentence embedding model, LaBSE \citep{Feng2020LanguageagnosticBS}, and show that the tuning improves the performance of parallel sentence matching for low-resource languages under-supported by the model.

Finally, we analyze the EASE model by studying ablated models and the multilingual properties of the sentence embeddings to shed light on the source of the improvement in the model.

\section{Related Work}

\subsection{Sentence Embeddings}
Sentence embeddings, which represent the meaning of sentences in the form of a dense vector, have been actively studied.
One of the earliest methods is Paragraph Vector \citep{Le2014Paragraph} in which sentence embeddings are trained to predict words within the text.
Subsequently, various kinds of training tasks have been explored including reconstructing or predicting adjacent sentences \citep{NIPS2015_f442d33f,DBLP:conf/iclr/LogeswaranL18} and solving a natural language inference (NLI) task \citep{conneau-etal-2017-supervised}.

Recently, with the advent of general-purpose pretrained language models such as BERT \citep{devlin-etal-2019-bert}, it has become increasingly common to fine-tune pretrained models to produce high-quality sentence embeddings, revisiting the aforementioned supervision signals \citep{reimers-gurevych-2019-sentence, yang-etal-2021-universal}, and using self-supervised objectives based on contrastive learning (CL).
In this paper, we present a CL objective with entity-based supervision.
We train our EASE model with entity CL together with self-supervised CL with dropout noise and show that the entity CL improves the quality of sentence embeddings.

\paragraph{Contrastive learning}
The basic idea of contrastive representation learning is to pull semantically similar samples close and keep dissimilar samples apart~\cite{1640964}.
CL for sentence embeddings can be classified by the type of positive pairs used.
As representative examples, several methods use entailment pairs as positive pairs in NLI datasets~\cite{gao-etal-2021-simcse, zhang-etal-2021-pairwise}.
To alleviate the need for an annotated dataset, self-supervised approaches are also being actively studied.
Typical self-supervised methods involve generating positive pairs using data augmentation techniques, including discrete operations such as word deletion and shuffling~\cite{yan-etal-2021-consert, meng2021cocolm}, back-translation~\cite{fang2020cert}, intermediate BERT hidden representations~\cite{kim-etal-2021-self}, and dropout noise within transformer layers~\cite{gao-etal-2021-simcse, liu-etal-2021-fast}.
Contrastive tension (CT)-BERT~\cite{DBLP:conf/iclr/CarlssonGGHS21} regards as positive pairs the outputs of the same sentence from two individual encoders.
DeCLUTR~\cite{giorgi-etal-2021-declutr} uses different spans of the same document.
In contrast to these methods that perform CL between sentences, our method performs CL between sentences and their associated entities.

\paragraph{Multilingual sentence embeddings}
Another yet closely related line of research is focused on learning multilingual sentence embeddings, which capture semantics across multiple languages.
Early competitive methods typically utilize the sequence-to-sequence objective with parallel corpora to learn multilingual sentence embeddings \citep{schwenk-douze-2017-learning,artetxe-schwenk-2019-massively}; recently fine-tuned multilingual pretrained models have achieved state-of-the-art performance \citep{Feng2020LanguageagnosticBS,goswami-etal-2021-cross}.
However, one drawback of such approaches is that, to achieve strong results for a particular language pair, they need rich parallel or semantically related sentence pairs, which are not necessarily easy to obtain.
In this work, we explore the utility of Wikipedia entity annotations, which are aligned across languages and already available in over 300 languages.
We also show that the entity CL in a multilingual scenario effectively improves the alignment of sentence embeddings between English and low-resource languages not well supported in an existing multilingual model.

\subsection{Learning Representations Using Entity-based Supervision}
Entities have been conventionally used to model text semantics~\cite{Gabrilovich2007ESA, DBLP:conf/aaai/GabrilovichM06}.
Several recently proposed methods learn text representations based on entity-based supervision by predicting entities from their relevant text~\cite{yamada-etal-2017-learning} or entity-masked sentences~\cite{Ling2020LearningCE}.
In the proposed EASE model, the existing self-supervised CL method based on BERT~\cite{gao-etal-2021-simcse} is extended using entity-based supervision with carefully designed hard negatives.
Moreover, it is applied to the multilingual setting by leveraging the language-agnostic nature of entities.

\section{Model and Training Data}
\label{seq: training_data}
In this section, we describe the components of our learning method for sentence embeddings, EASE, which is trained using entity hyperlink annotations available in Wikipedia.

\subsection{Contrastive Learning with Entities}
\label{subseq: entity_cl}
Given pairs of a sentence and a semantically related entity (positive entity) $\mathcal{D} = \{(s_i, e_i) \}_{i=1}^m$, we train our model to predict the entity embedding $\bf{e_i}  \in \mathbb{R}^{\rm{d_{e}}}$ from the sentence embedding $\bf{s_i}  \in \mathbb{R}^{\rm{d_{s}}}$.
Following the contrastive framework in \citet{pmlr-v119-chen20j}, the training loss for $(s_i, e_i)$ with a minibatch of $N$ pairs is:

\begin{equation}
        l^e_i = -\log \frac{e^{\rm{sim}(\bf{s}_i, W\bf{e}_i)/\tau}}{\sum_{j=1}^N e^{\rm{sim}(\bf{s}_i, W\bf{e}_j)/\tau}},
    \label{eq:entity_objective}
\end{equation}

\noindent where $\mathbf{W} \in \mathbb{R}^{\rm{d_{e}} \times \rm{d_{s}}}$ is a learnable matrix weight, $\tau$ is a temperature hyperparameter, and $\rm{sim(\cdot)}$ is the cosine similarity $\frac{\bf{s}_1^\top \bf{s}_2}{\Vert \bf{s}_1\Vert \cdot \Vert \bf{s}_2\Vert}$.

\paragraph{Data}
We construct the sentence-entity paired datasets from the January 2019 version of Wikipedia dump.
We split text in the articles into sentences using \texttt{polyglot}.\footnote{\url{https://polyglot.readthedocs.io/en/latest/Tokenization.html}}
For each sentence, we extract the hyperlink entities as semantically related entities.\footnote{In a preliminary experiment, we also tried constructing entity-sentence paired data from entities and the first sentence on their page, and found that the current approach performs better.}
Each entity forms a training instance $(s_i, e_i) $ for the sentence.
We restrict the entities to those that appear more than ten times as hyperlinks in the training corpus.
They are converted into Wikidata entities, which are shared across languages, using inter-language links obtained from the March 2020 version of the Wikidata dump.\footnote{\url{https://en.wikipedia.org/wiki/Help:Interlanguage_links}}

\subsection{Hard Negative Entities}
The introduction of hard negatives (data that are difficult to distinguish from an anchor point) has been reported to be effective in improving CL models~\cite{gao-etal-2021-simcse, DBLP:conf/iclr/RobinsonCSJ21}.
We introduce a hard negative mining technique that finds negative entities similar to the positive entity but yet unrelated to the sentence.

Specifically, for each positive entity, we collect hard negative entity candidates that satisfy the following two conditions:
(1) entities with the same type as the positive entity.
Entity types are defined as the entities in the ``instance of'' relation on Wikidata, following the work of \citet{DBLP:conf/iclr/XiongDWS20}.
If there are more than one appropriate type, we randomly choose one;
(2) entities that do not appear on the same Wikipedia page.
Our assumption here is that entities on the same page are topically related to the positive entity and thus are not appropriate for negative data.
Finally, we randomly choose one of the candidates to construct hard negative training data.
For example, the ``Studio Ghibli'' entity has the type ``animation studio'' and one of the hard negative entity candidates is ``Walt Disney Animation Studios''.

Given datasets with hard negative entities~$\mathcal{D} = \{(s_i, e_i, e^{-}_i) \}_{i=1}^m$, the loss function is

\footnotesize
\begin{equation}
        l^e_i = -\log \frac{e^{\rm{sim}(\bf{h}_i, W\bf{e}_i)/\tau}}{\sum_{j=1}^N (e^{\rm{sim}(\bf{h}_i, W\bf{e}_j)/\tau} + e^{\rm{sim}(\bf{h}_i, W\bf{e^{-}_j})/\tau})}.
    \label{eq:ease_entity_objective}
\end{equation}
\normalsize

\subsection{Pretrained Entity Embeddings}
\label{subseq: entity_embs}
We initialize entity embeddings using English entity embeddings pretrained on Wikipedia.
These embeddings are trained using the open-source \texttt{Wikipedia2Vec} tool~\cite{yamada-etal-2020-wikipedia2vec} and the January 2019 English Wikipedia dump.
The vector dimension is set to 768, which is the same as those of the hidden representations of the base pretrained models, and the other hyperparameters to their default values.
The parameters of the entity embedding matrix are updated during the training process.

\subsection{Self-supervised Contrastive Learning with Dropout Noise}
Self-supervised CL with dropout noise, which inputs a sentence and predicts itself using dropout as noise, is an effective method for learning sentence embeddings in an unsupervised way~\cite{liu-etal-2021-fast, gao-etal-2021-simcse}.
We combine this method with our entity CL.

Given two embeddings with different dropout masks $\bf{s}_i, \bf{s}^+_i$,
the training loss of self-supervised CL $l^s_i$ is defined by

\begin{equation}
        l^s_i = -\log \frac{e^{\rm{sim}(\bf{s}_i, \bf{s}^+_i)/\tau}}{\sum_{j=1}^N e^{\rm{sim}(\bf{s}_i, \bf{s}_j^+)/\tau}}.
    \label{eq:ease_cl_objective}
\end{equation}

\noindent In summary, our total loss is

\begin{equation}
    l^{ease}_i = \lambda l^e_i + l^s_i,
    \label{eq:ease_full_objective}
\end{equation}

\noindent where $l^e$ and $l^s$ are defined in Equations (\ref{eq:ease_entity_objective}) and (\ref{eq:ease_cl_objective}) respectively,
and $\lambda$ denotes a hyperparameter that defines the balance between the entity CL and self-supervised CL with dropout noise.
The details on the hyperparameters of the models can be found in \Appendix{appendix:training_details}.

\section{Experiment: Monolingual}

We first evaluate EASE in monolingual settings. We fine-tune monolingual pre-trained language models using only English Wikipedia data.

\subsection{Setup}
We use one million pairs sampled from the English entity-sentence pairs described in Section \ref{seq: training_data} as training data.
In this setting, we train sentence embedding models from pre-trained checkpoints of BERT~\cite{devlin-etal-2019-bert} or RoBERTa~\cite{liu2019roberta} and take the [CLS] representation as the sentence embedding.
We add a linear layer after the output sentence embeddings only during training, as in \citet{gao-etal-2021-simcse}.

We compare our method with unsupervised sentence embedding methods including average GloVe embeddings~\cite{pennington-etal-2014-glove}, average embeddings of vanilla BERT or RoBERTa, and previous state-of-the-art approaches such as SimCSE~\cite{gao-etal-2021-simcse}, CT~\cite{DBLP:conf/iclr/CarlssonGGHS21}, and DeCLUTR~\cite{giorgi-etal-2021-declutr}.

We evaluate sentence embeddings using two tasks: STS and STC.
These tasks are supposed to measure the degree of sentence embeddings capturing fine-grained and broad semantic structures.

\begin{table}[t]
	\begin{center}
	\scalebox{0.8}{ 
\begin{tabular}{l|c|c}
\toprule
Model & 7 STS avg. & 8 STC avg.  \\ \midrule
GloVe embedding (avg.) & $61.3^\dagger $ & 56.4  \\
BERT (avg.)  & 52.6 & 50.9  \\
CT-BERT\ba{}  & 72.1 & 61.6 \\
SimCSE-BERT\ba{}  & 76.3 & 57.1 \\
\textbf{EASE-BERT\ba{}} & \textbf{77.0} & \textbf{63.1} \\
\hline
RoBERTa (avg.)  & 53.5 & 40.9  \\
DeCLUTR-RoBERTa\ba{} & 70.0 & \textbf{60.0}  \\
SimCSE-RoBERTa\ba{}  & 76.6 & 57.4  \\
\textbf{EASE-RoBERTa\ba{}} & \textbf{76.8} & 58.6  \\
\bottomrule
\end{tabular}
}
\end{center}
\caption{Sentence embedding performance on seven monolingual STS tasks (Spearman’s correlation) and eight monolingual STC tasks (clustering accuracy). The highest values among the models with the same pre-trained encoder are in bold. $\dagger$: results from \citet{reimers-gurevych-2019-sentence}; all other results are reproduced or reevaluated by us using published checkpoints. The complete results are available in Appendix~\ref{app: monolingul_full_results}.}
\label{tab: mono-sts-stc}
\end{table}

\begin{table*}[t]
	\begin{center}
	\scalebox{0.70}{ 
	\begin{tabular}{l|ccccccccccc}
		\toprule
		\textbf{Model} & \textbf{EN-EN}  & \textbf{AR-AR}  & \textbf{ES-ES}  & \textbf{EN-AR} & \textbf{EN-DE} & \textbf{EN-TR} & \textbf{EN-ES} & \textbf{EN-FR} & \textbf{EN-IT} & \textbf{EN-NL} & \textbf{Avg.} \\ \midrule
		mBERT\ba{} (avg.)  & 54.4& 50.9 & 56.7 & 18.7 & 33.9 & 16.0 & 21.5 & 33.0 & 34.0 & 35.3 & 35.4 \\
		SimCSE-mBERT\ba{} & 78.3&62.5&76.7&26.2&55.6&23.8&37.9&48.1&49.6&50.3&50.9  \\
		\textbf{EASE-mBERT\ba{}} & \textbf{79.3} & \textbf{62.8} & \textbf{79.4}& \textbf{31.6}& \textbf{59.8} & \textbf{26.4} & \textbf{53.7} & \textbf{59.2} & \textbf{59.4} & \textbf{60.7} & \textbf{57.2} \\ \hline
		XLM-R\ba{} (avg.) & 52.2 & 25.5 & 49.6 & 15.7 & 21.3 & 12.1 & 10.6 & 16.6 & 22.9& 23.9& 25.0 \\
		SimCSE-XLM-R\ba{} & 77.9&63.4&\textbf{80.6}&\textbf{36.3}&56.2&28.9&38.9&\textbf{51.8}&52.6&54.2&54.1  \\
		\textbf{EASE-XLM-R\ba{}} & \textbf{80.6} & \textbf{65.3} & 80.4 & 34.2 &\textbf{59.1} & \textbf{37.6} & \textbf{46.5} & 51.2 & \textbf{56.6} & \textbf{59.5} & \textbf{57.1}\\
		\bottomrule
	\end{tabular}}
	\end{center}
	\caption{Spearman's correlation for multilingual semantic textual similarity on extended version of STS 2017 dataset.}
	\label{tab: msts}
\end{table*}

\begin{table*}[t]
	\begin{center}
	\scalebox{0.70}{ 
	\begin{tabular}{l|ccccccccccccccccc}
		\toprule
		\textbf{Model} &  ar & ca & cs  & de & en & eo & es & fa & fr & ja & ko & pl & pt  & ru & sv & tr & Avg. \\ \hline
		mBERT\ba{} (avg.) & 27.0&27.2&\textbf{44.3}&36.2&37.9&25.6&\textbf{41.1}&35.0&25.9&44.2&31.0&35.0&30.1&23.4&28.9&34.9&33.0\\
		SimCSE-mBERT\ba{} & 30.1&26.9&41.3&32.5&37.3&27.2&36.2&\textbf{36.9}&29.0&48.9&33.9&37.6&37.9&\textbf{27.1}&26.9&35.3&34.1 \\
		\textbf{EASE-mBERT\ba{}}& \textbf{31.9}&\textbf{29.6}&38.8&\textbf{38.5}&\textbf{30.2}&\textbf{34.5}&37.2&36.7&\textbf{30.4}&\textbf{49.3}&\textbf{36.2}&\textbf{40.0}&\textbf{41.0}&27.0&\textbf{30.5}&\textbf{44.7}&\textbf{36.0} \\ \hline
		XLM-R\ba{} (avg.)  & \textbf{26.0}&24.7&28.2&29.4&23.0&23.5&22.1&36.6&23.6&38.8&22.0&24.2&32.8&18.0&\textbf{33.2}&26.0&27.0 \\
		SimCSE-XLM-R\ba{}& 24.6&26.3&34.6&28.6&33.4&31.7&32.9&35.9&29.1&41.1&31.1&33.1&30.0&26.0&32.9&37.2&31.8 \\
		\textbf{EASE-XLM-R\ba{}} & 25.3&\textbf{26.7}&\textbf{43.2}&\textbf{37.0}&\textbf{34.9}&\textbf{34.2}&\textbf{37.2}&\textbf{42.4}&\textbf{32.0}&\textbf{46.0}&\textbf{32.8}&\textbf{41.6}&\textbf{33.4}&\textbf{31.3}&27.2&\textbf{41.8}&\textbf{35.4}\\
		\bottomrule

	\end{tabular}}
	\end{center}
	\caption{Clustering accuracy for multilingual short text clustering on MewsC-16 dataset.}
	\label{tab: mstc}
\end{table*}

\paragraph{Semantic textual similarity}

\label{para: mono-sts}
STS is a measure of the capability of capturing graded similarity of sentences.
We use seven monolingual STS tasks: STS 2012-2016~\cite{agirre-etal-2012-semeval,agirre-etal-2013-sem, agirre-etal-2014-semeval,agirre-etal-2015-semeval,agirre-etal-2016-semeval}, STS Benchmark~\cite{cer-etal-2017-semeval}, and SICK-Relatedness~\cite{marelli-etal-2014-sick}.
Following the settings of \citet{reimers-gurevych-2019-sentence}, we calculate Spearman's rank correlation coefficient between the cosine similarity of the sentence embeddings and the ground truth similarity scores.

\paragraph{Short text clustering}

\label{para: mono-stc}
Another important aspect of sentence embeddings is the ability to capture categorical semantic structure, i.e., to map sentences from the same categories close together and those from different categories far apart~\cite{zhang-etal-2021-pairwise}.
We also evaluate sentence embeddings using eight benchmark datasets for STC~\cite{zhang-etal-2021-pairwise} to investigate how well our method can encode high-level categorical structures into sentence embeddings.
These datasets contain short sentences, ranging from 6 to 28 average words in length, from a variety of domains such as news, biomedical, and social network service (Twitter).
We cluster the sentence embeddings using K-Means~\cite{MacQueen1967} and compute the clustering accuracy using the Hungarian algorithm~\cite{Munkres1957Assignment} averaged over three independent runs.

\subsection{Results}

Table~\ref{tab: mono-sts-stc} shows the evaluation results for the seven STS and eight STC tasks.
Overall, our EASE methods significantly improve the performance of the base models (i.e., BERT and RoBERTa), and on average outperform the previous state-of-the-art methods on all tasks except STC with the RoBERTa backbone.
The most significant improvement is observed for EASE-BERT, with an average
improvement of $61.6\%$ to $63.1\%$ over the previous best result for STC tasks.
These results suggest that EASE is able to measure the semantic similarity between sentences, and simultaneously excel at capturing high-level categorical semantic structure.

\section{Experiment: Multilingual}
\label{sec: multilingual_experiment}
To further explore the advantage of entity annotations as cross-lingual alignment supervision, we test EASE in multilingual settings: we fine-tune multilingual pre-trained language models using Wikipedia data in multiple languages.

\subsection{Setup}

We sample 50,000 pairs for each language and use them together as training data from the entity-sentence paired data in 18 languages.\footnote{We chose 18 languages (ar, ca, cs, de, en, eo, es, fa, fr, it, ja, ko, nl, pl, pt, ru, sv, tr) present in both the MewsC-16 dataset (see Section \ref{subseq: MewsC-16}) and the extended version of STS 2017.}
As our primary baseline model, we use a SimCSE model trained using the same multilingual data as EASE (i.e., sentences in entity-sentence paired data).\footnote{We use the same entity embeddings trained on English Wikipedia as those of the monolingual settings (Section \ref{subseq: entity_embs}). Note that entities in all languages are converted to Wikidata entities that are shared across languages using inter-language links as described in Section \ref{subseq: entity_cl}.}
In this setting, we start fine-tuning from pre-trained checkpoints of mBERT or XLM-R~\cite{conneau-etal-2020-unsupervised} and take mean pooling to obtain sentence embeddings for both training and evaluation on both EASE and SimCSE.
We also tested other pooling methods, but mean pooling was the best in this experiment for both models (Appendix~\ref{app: pooling_method}).

\begin{table*}[htbp]
	\begin{center}
	\scalebox{0.70}{ 
	\begin{tabular}{l|ccccccccccccccccc}
		\toprule
		\textbf{Model} &  ar & ca & cs  & de & eo & es & fr & it & ja & ko & nl & pl & pt  & ru & sv & tr & Avg. \\ \midrule
		mBERT\ba{} (avg.) &20.6&49.2&32.8&62.8&12.2&57.7&55.6&50.8&38.6&33.1&54.8&40.2&58.5&51.4&45.8&30.1&43.4 \\
		SimCSE-mBERT\ba{} & 16.4&51.5&30.7&57.0&18.2&54.8&54.5&49.9&39.6&28.1&52.7&37.9&53.6&46.8&45.5&25.0&41.4 \\
		\textbf{EASE-mBERT\ba{}} &\textbf{32.1}&\textbf{66.5}&\textbf{47.7}&\textbf{74.2}&\textbf{26.1}&\textbf{70.1}&\textbf{66.7}&\textbf{65.3}&\textbf{59.2}&\textbf{46.8}&\textbf{69.2}&\textbf{55.4}&\textbf{69.1}&\textbf{64.4}&\textbf{59.4}&\textbf{38.1}&\textbf{56.9}\\ \hline
		XLM-R\ba{} (avg.)  & 10.3&15.3&16.5&49.6&7.5&36.4&30.8&25.6&15.0&19.3&45.2&24.1&42.0&37.4&42.8&17.9&27.2 \\
		SimCSE-XLM-R\ba{} & 38.4&57.6&55.7&80.6&46.0&68.9&70.4&66.4&60.0&54.1&73.1&65.3&75.1&71.1&76.7&56.4&63.5 \\
		\textbf{EASE-XLM-R\ba{}} &\textbf{42.6}&\textbf{65.1}&\textbf{63.8}&\textbf{87.2}&\textbf{56.1}&\textbf{75.9}&\textbf{74.1}&\textbf{70.8}&\textbf{68.2}&\textbf{60.5}&\textbf{77.9}&\textbf{71.9}&\textbf{80.6}&\textbf{76.5}&\textbf{79.2}&\textbf{60.9}&\textbf{69.4}\\
		\bottomrule
	\end{tabular}}
	\end{center}
	\caption{Accuracy on Tatoeba dataset averaged over forward and backward
directions (en to target language and vice-versa).}
	\label{tab: tatoeba}
\end{table*}

\begin{table*}[ht]
  \begin{minipage}[t]{.3\textwidth}
      \begin{center}
    \scalebox{0.7}{ 
	\begin{tabular}{l|cc}
		\toprule
		\textbf{Model} &  Avg. \\ \midrule
		mBERT\ba{} (avg.) & 17.3 \\
		SimCSE-mBERT\ba{} & 16.8 \\
		\textbf{EASE-mBERT\ba{}} &\textbf{25.4} \\ \hline
		XLM-R\ba{} (avg.)  & 9.4 \\
		SimCSE-XLM-R\ba{} & 28.5 \\
		\textbf{EASE-XLM-R\ba{}} &\textbf{32.1} \\
		\bottomrule
	\end{tabular}
	}
	\caption{Average accuracy for 94 languages not included in EASE training on Tatoeba.}
	\label{tab: tatoeba_unseen}
    \end{center}
  \end{minipage}
  \hfill
  \begin{minipage}[t]{.65\textwidth}

      \begin{center}
    \scalebox{0.7}{ 
	\begin{tabular}{l|ccccccccc}
		\hline
		\textbf{Model} & en (dev) & de & es  & fr & it & ja & ru & zh & Avg. \\ \hline
		mBERT\ba{} (avg.) & \textbf{89.5}&68.0&68.1&70.6&62.7&61.2&61.5&69.6&65.9 \\
		SimCSE-mBERT\ba{} & 88.4&62.3&\textbf{73.2}&78.2&64.3&\textbf{63.7}&61.3&\textbf{75.0}&68.3 \\
		\textbf{EASE-mBERT\ba{}} & 89.0&\textbf{69.9}&69.2&\textbf{80.1}&\textbf{66.8}&62.8&\textbf{64.4}&73.2&\textbf{69.5} \\ \hline
		XLM-R\ba{} (avg.)  & \textbf{90.9} &\textbf{82.7}&\textbf{79.8}&72.1&72.5&71.1&69.6&71.4&74.2 \\
		SimCSE-XLM-R\ba{}& 90.7&74.9&74.1&81.5&70.3&71.7&70.1&76.6&74.2 \\
		\textbf{EASE-XLM-R\ba{}} &90.6&77.9&75.6&\textbf{83.9}&\textbf{72.6}&\textbf{72.8}&\textbf{71.1}&\textbf{81.6}&\textbf{76.5}\\
		\hline
	\end{tabular}
	}
	\end{center}
	\caption{Classification accuracy for zero-shot cross-lingual text classification on MLDoc dataset.}
	\label{tab: mldoc}
  \end{minipage}
\end{table*}

\subsection{Multilingual STS and STC}

We evaluate our method using the extended version of the STS 2017 dataset~\cite{reimers-gurevych-2020-making}, which contains annotated sentences for ten language pairs: EN-EN, AR-AR, ES-ES, EN-AR, EN-DE, EN-TR, EN-ES, EN-FR, EN-IT, and EN-NL.
We compute Spearman's rank correlation as in Section~\ref{para: mono-sts}.
We also conduct experiments on our newly introduced multilingual STC dataset described as follows:

\paragraph{MewsC-16}
\label{subseq: MewsC-16}
To evaluate the ability of sentence embeddings to encode high-level categorical concepts in a multilingual setting, we constructed MewsC-16 (\textbf{M}ultilingual Short Text \textbf{C}lustering Dataset for N\textbf{ews} in \textbf{16} languages) from Wikinews.\footnote{\url{https://en.wikinews.org/wiki/Main_Page}}
MewsC-16 contains topic sentences from Wikinews articles in 13 categories and 16 languages.
More detailed information is available in Appendix~\ref{app: MewsC-16}.
We perform clustering and compute the accuracy for each language as in Section~\ref{para: mono-stc}.

Tables~\ref{tab: msts} and \ref{tab: mstc} show the results of our multilingual STS and STC experiments.
Overall, EASE substantially outperforms the corresponding base models (i.e., mBERT and XLM-R) on both tasks.
Similar to the results for the monolingual setting, the average performance of EASE exceeds that of SimCSE for multilingual STC tasks with an improvement of $34.1\%$ to $36.0\%$ for mBERT and $31.8\%$ to $35.4\%$ for XLM-R.
This result suggests that even in a multilingual setting, EASE can encode high-level categorical semantic structures into sentence embeddings.
Moreover, EASE significantly outperforms SimCSE in multilingual STS tasks
Specifically, the score of EASE-mBERT is better than that of SimCSE-mBERT ($50.9$ vs $57.2$), and that of EASE-XLM-R is better than that of SimCSE-XLM-R ($54.1$ vs $57.1$).
This improvement is much more significant than the monolingual setting (Table~\ref{tab: mono-sts-stc}), where the improvement is less than one point.
This indicates the effectiveness of language-independent entities as cross-lingual alignment supervision in learning multilingual sentence embeddings.

\subsection{Cross-lingual Parallel Matching}
\label{subsec:tatoeba}
We evaluate EASE on the Tatoeba dataset \citep{artetxe-schwenk-2019-massively}
to assess more directly its ability to capture cross-lingual semantics.
This task is to retrieve the correct target sentence for each query sentence, given a set of parallel sentences.
We perform the retrieval using the cosine similarity scores of the sentence embeddings.
For each language-pair dataset, we compute the retrieval accuracy averaged over the forward and backward directions (English to the target language and vice-versa).

Table~\ref{tab: tatoeba} shows the evaluation results for the languages in the CL training data.
EASE significantly outperforms the corresponding base models and SimCSE for all languages.
Notably, the mean performance of EASE-mBERT is better than that of vanilla mBERT by $13.5$ percentage points.
This indicates that EASE can capture cross-lingual semantics owing to the cross-lingual supervision of entity annotations, which aligns semantically similar sentences across languages.
One interesting observation is that the performance of SimCSE-mBERT is worse than that of vanilla mBERT.
We conjecture that this is because the SimCSE model is trained using only the positive sentence pairs within the same language, which sometimes leads to less language-neutral representations.

To further explore the cross-lingual ability of EASE, we evaluate it on languages not included in the EASE training set (Table~\ref{tab: tatoeba_unseen}).
The results show that EASE performs robustly on these languages as well, which suggests that, in EASE, the cross-lingual alignment effect propagates to other languages not used in additional training with EASE \citep{kvapilikova-etal-2020-unsupervised}.


\begin{table*}[t]
\begin{center}
\scalebox{0.85}{ 
\begin{tabular}{l|cccc}
\toprule
\multirow{3}{*}{Setting} & EASE-BERT\ba & EASE-RoBERTa\ba &EASE-mBERT\ba & EASE-XLM-R\ba\\
 & STS avg. & STS avg. & mSTS avg. & mSTS avg. \\
\midrule

Full model  & \textbf{76.9} & \textbf{76.8} & \textbf{57.2}  & \textbf{57.1} \\
\qquad w/o self-supervised CL & 65.3 & 66.1  & 49.3  & 53.1   \\
\qquad w/o hard negative & 75.3 & 76.1 & 53.8 &  52.7  \\
\qquad w/o Wikipedia2Vec & 73.8& 76.3 & 52.1 & 54.3  \\
\qquad w/o all (vanilla model)  & 31.4 & 43.6 & 35.4  & 25.0 \\
\bottomrule
\end{tabular}
}
\end{center}
\caption{Results of ablation study.}
\label{tab: ablation}
\end{table*}

\subsection{Cross-lingual Zero-shot Transfer}
We further evaluate our sentence embeddings on a downstream task in which sentence embeddings are used as input features, especially in the cross-lingual zero-shot transfer setting.
For evaluation in this setting, we use MLDoc~\cite{schwenk-li-2018-corpus}, a cross-lingual document classification dataset that classifies news articles in eight languages into four categories.
We train a linear classifier using sentence embeddings as input features on the English training data, and evaluate the resulting classifier in the remaining languages.
To directly evaluate the ability of the resulting sentence embeddings, we do not update the parameters of the sentence encoder but only train the linear classifier in this setting.
The detailed settings are shown in Appendix~\ref{app: mldoc}.

As shown in Table~\ref{tab: mldoc}, our EASE models achieve the best average performance on both back-bones, suggesting that multilingual embeddings learned with the CL are also effective in the cross-lingual transfer setting.

\section{Case Study: Fine-tuning Supervised Model with EASE}
\label{sec:case_study}

Existing multilingual sentence representation models trained on a large parallel corpus do not always perform well, especially for languages that are not included in the training data.
In contrast, EASE requires only the Wikipedia text corpus, which is available in more than 300 languages.\footnote{\url{https://meta.wikimedia.org/wiki/List_of_Wikipedias}}
Thus, one possible use case for EASE would be to complement the performance of existing models in low-resource languages by exploiting the Wikipedia data in those languages.

To test this possibility, we fine-tune LaBSE \citep{Feng2020LanguageagnosticBS}, which is trained on both monolingual and bilingual data in 109 languages, with our EASE framework in five low-resource languages (kab, pam, cor, tr, mhr).
These languages are not present in the original training corpus, so the model performed particularly poorly on these languages.
We fine-tune the model using 5,000 pairs each from English and the corresponding language data.

\begin{figure}[t]
  \centering
  \includegraphics[width=7.5cm]{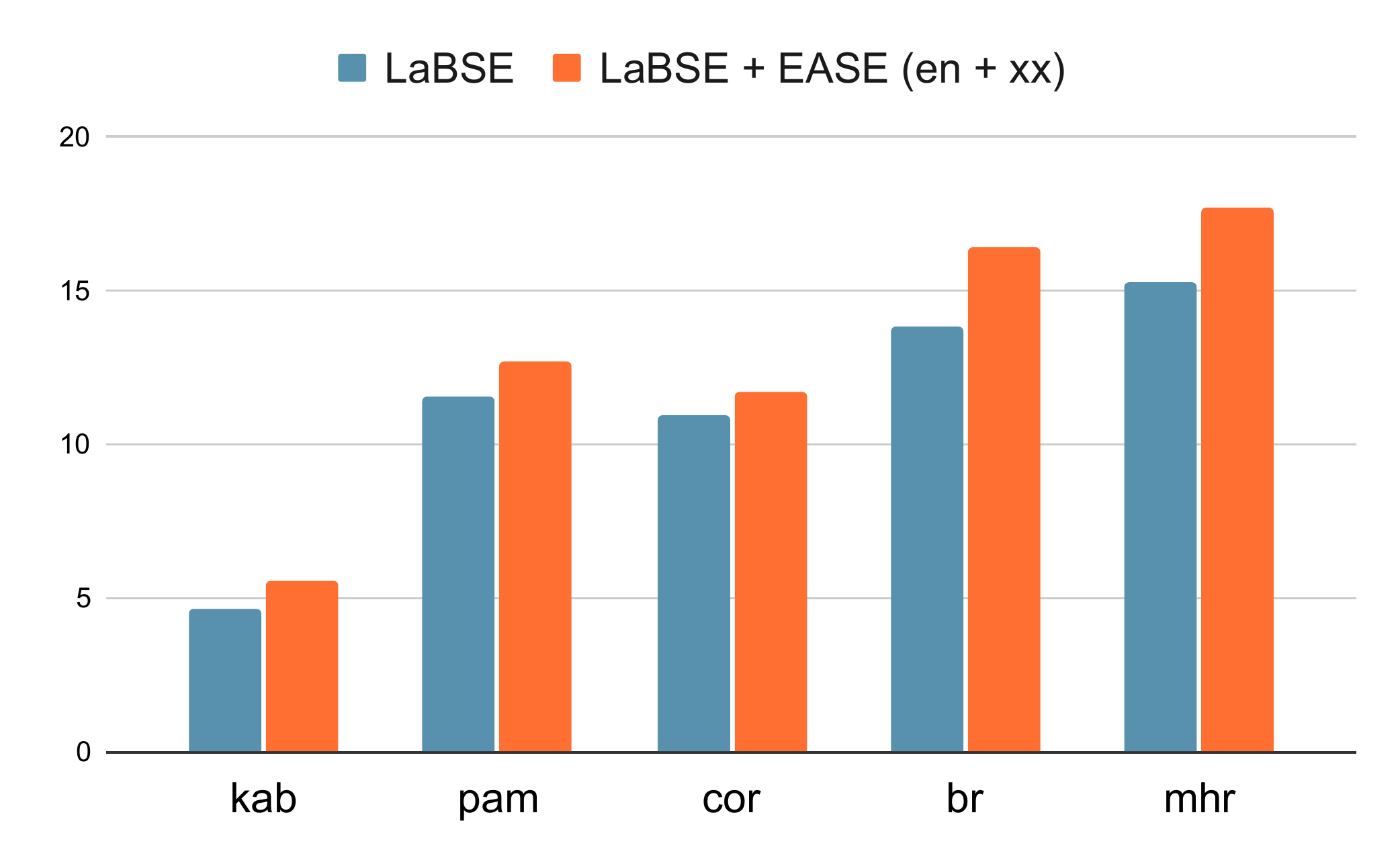}
  \caption{Results of fine-tuning LaBSE with EASE framework on Tatoeba dataset.}
  \label{fg: tune-labse}
\end{figure}

As shown in Figure~\ref{fg: tune-labse}, EASE improves the performance of LaBSE across all target languages, which is an intriguing result considering that LaBSE has already been trained on about six billion parallel corpora.
These results suggest the potential benefit of combining EASE with other models using parallel corpora, especially for languages without or with only a few parallel corpora.

\section{Analysis}

\subsection{Ablation Study}

We conduct ablation experiments to better understand how each component of EASE contributes to its performance.
We measure the performance of the models using monolingual STS in the monolingual setting and multilingual STS in the multilingual setting, without one of the following components: the self-supervised CL loss, hard negatives, and Wikipedia2Vec initialization (Table~\ref{tab: ablation}).
As a result, we find all of the components to make an important contribution to the performance.

It is worth mentioning that entity CL alone (i.e., w/o self-supervised CL) also improves the baseline performance significantly.
The performance contributions in the multilingual setting are particularly significant ($49.3$ for mBERT and $53.1$ for XLM-R) and comparable to those for the SimCSE models.
These results suggest that CL with entities by itself is effective in learning multilingual sentence embeddings.

\subsection{Alignment and Uniformity}
\begin{figure}[t]
  \centering
  \includegraphics[width=7.5cm]{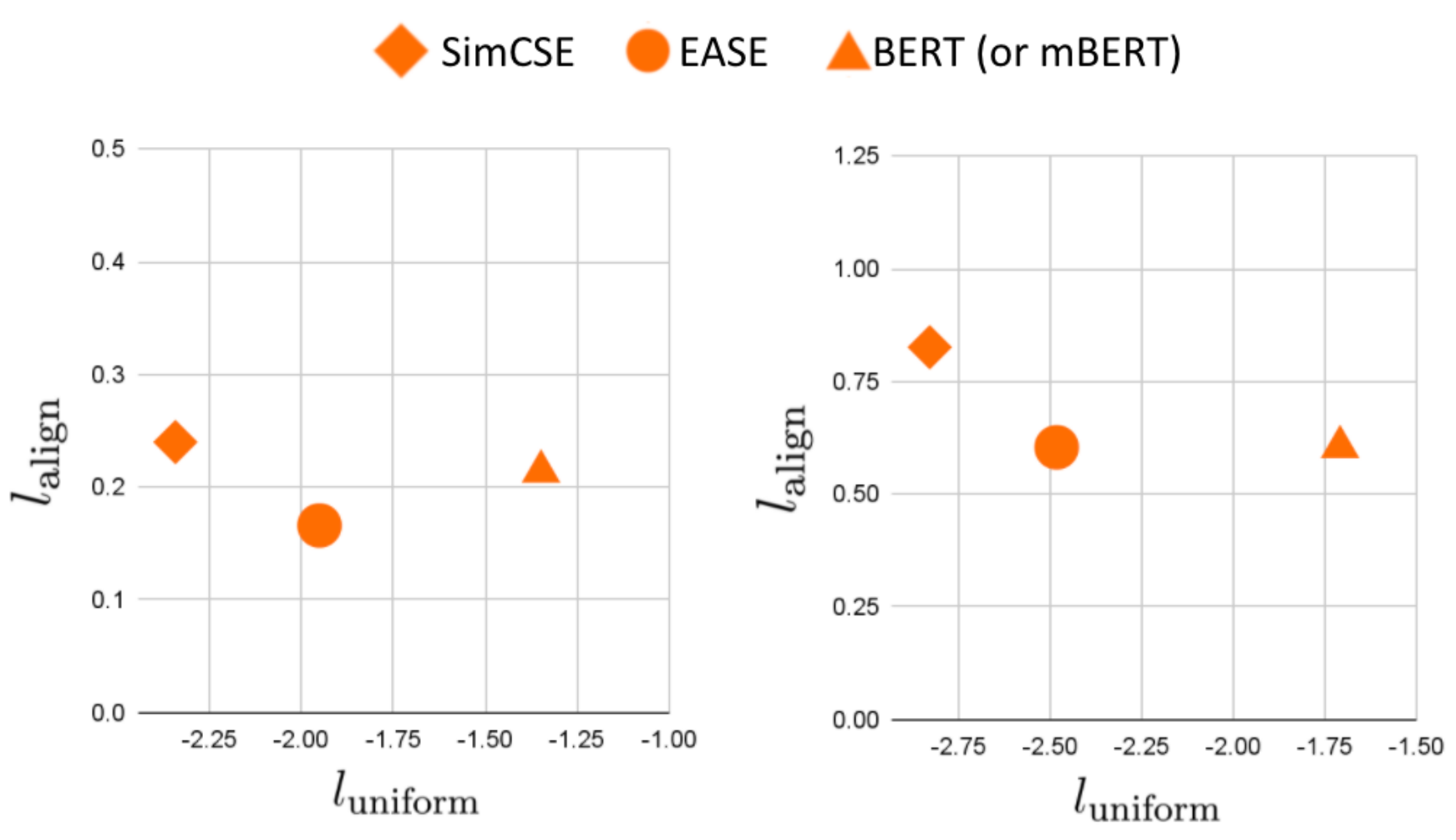}
  \caption{$l_{\rm{align}}-l_{\rm{uniform}}$ plot of BERT-based (or mBERT-based) models in monolingual (left) and multilingual (right) settings.}
  \label{fg: alignment_uniformity}
  \vspace{-2mm}
\end{figure}

To further understand the source of the performance improvement with EASE,
we evaluate two key properties to measure the quality of the representations obtained from contrastive learning~\cite{DBLP:conf/icml/0001I20}:
\textit{alignment} measures the closeness of representations between positive pairs;
\textit{uniformity} measures how well the representations are uniformly distributed.
We let $f(x)$ denote the normalized representation of $x$,
and compute the two measures using
  \vspace{-2mm}
\begin{equation}
    \label{eq:alignment}
    l_{\rm{align}}
    \triangleq \underset{(x, x^+)\sim p_{\rm{pos}}}{\mathbb{E}} \Vert f(x) - f(x^+) \Vert^2,
\end{equation}

\begin{equation}
    \label{eq:uniformity}
        l_{\rm{uniform}}\triangleq\log \underset{~~~x, y\stackrel{i.i.d.}{\sim} p_{\rm{data}}}{\mathbb{E}}   e^{-2\Vert f(x)-f(y) \Vert^2},
\end{equation}

\noindent where $p_{\rm{pos}}$ denotes positive pairs and $p_{\rm{data}}$ denotes the entire data distribution.
We compute these metrics using BERT-based models on the STS-B development set data.
For investigation in the multilingual setting, we compute them using mBERT-based models on the multilingual STS data used in the experiment Section~\ref{sec: multilingual_experiment}.
We compute the averages of alignment and uniformity for each language pair.
For each setting, we take STS pairs with a score higher than 4 in the 0-to-5 scale as $p_{\rm{pos}}$ and all STS sentences as $p_{\rm{data}}$.

As shown in Figure \ref{fg: alignment_uniformity}, the trends are similar in both settings:
(1) both EASE and SimCSE significantly improve uniformity compared with that for the vanilla model;
(2) EASE is inferior to SimCSE in terms of uniformity and superior in terms of alignment.
This result suggests that entity CL does not have the effect of biasing embeddings towards a more uniform distribution. Instead, it has the effect of aligning semantically similar samples, which leads to the improved performance of the resultant sentence embeddings.

\subsection{Qualitative analysis}

\begin{table*} [htp]
	\begin{center}
    \begin{subtable}[c]{1.0\textwidth}
        \centering
        \scalebox{0.9}{ 
        \begin{tabular}{p{15em}|p{15em}|c|c|cc} \hline
        \multicolumn{1}{c|}{Sentence1} & \multicolumn{1}{|c|}{Sentence2} & Gold & EASE & SimCSE \\ \hline
        i think you 're looking for mikey ( 1992 ) .&i think you 're looking for the movie
&\multirow{1}{*}[-1.5ex]{3.00} & \multirow{1}{*}[-1.5ex]{2.32} & \multirow{1}{*}[-1.5ex]{1.62} \\ \hline

the new york senator 's new book , " living history , " appears a certain bestseller . &hillary clinton , the new york senator and former first lady , has a book out monday titled living history .&
\multirow{1}{*}[-3ex]{3.20} & \multirow{1}{*}[-3ex]{3.57} & \multirow{1}{*}[-3ex]{1.94} \\ \hline

he was referring to john s. reed , the former citicorp chief executive who became interim chairman and chief executive of the exchange last sunday . &
next week , john s. reed , the former citicorp chief executive who sunday became interim chairman and chief executive of the exchange , will take up his position . &
\multirow{1}{*}[-5ex]{4.00} & \multirow{1}{*}[-5ex]{3.52} & \multirow{1}{*}[-5ex]{2.73} \\ \hline
 
        \end{tabular}
        }
        \caption{Improvement cases \label{tab: imp_cases}}
        \vspace{5mm}   
    \end{subtable}
\\
    \begin{subtable}[c]{1.0\textwidth}
        \centering
        \scalebox{0.9}{ 
        \begin{tabular}{p{15em}|p{15em}|c|c|cc} \hline
        \multicolumn{1}{c|}{Sentence1} & \multicolumn{1}{|c|}{Sentence2} & Gold & EASE & SimCSE \\ \hline
        it 's not a good idea . & it 's a good question . & 0.00 & 2.88 & 1.33 \\ \hline
suicide attack kills eight in baghdad & suicide attacks kill 24 people in baghdad &
\multirow{1}{*}[-1.5ex]{2.40} & \multirow{1}{*}[-1.5ex]{3.92} & \multirow{1}{*}[-1.5ex]{2.43} \\ \hline
the nasdaq composite index rose 19.67 , or 1.3 percent , to 1523.71 , its highest since june 18 . & the s and p 500 had climbed 16 percent since its march low and yesterday closed at its highest since dec. 2 .& \multirow{1}{*}[-3ex]{0.80}&\multirow{1}{*}[-3ex]{3.25}&\multirow{1}{*}[-3ex]{2.04} \\ \hline
        \end{tabular}
        }
        \caption{Deterioration cases \label{tab: det_cases}}
        \vspace{5mm}           
    \end{subtable}
    \caption{Comparison of STS Benchmark results by monolingual EASE and SimCSE.}
    \end{center}
\end{table*}

To investigate how EASE improves the quality of sentence embeddings,
we conduct qualitative analysis by comparing the results of EASE and SimCSE on STS Benchmark.
Table \ref{tab: imp_cases} shows typical cases of how EASE improves sentence embeddings.
We find that EASE embeddings are more robust to synonyms and grammatical differences
since they are more aware of the topic similarity between sentences,
resulting in more accurate score inference.
On the other hand, as shown in the deterioration cases in Table \ref{tab: det_cases},
EASE embeddings are sometimes overly sensitive to topical similarity,
making it difficult to capture the correct meaning of the whole sentence.

\section{Discussion and Conclusion}
Our experiments have demonstrated that entity supervision in EASE improves the quality of sentence embeddings both in the monolingual setting and, in particular, the multilingual setting.
As recent studies have shown, entity annotations can be used as \emph{anchors} to learn quality cross-lingual representations \citep{Calixto2021naacl,nishikawa2021multilingual,XLM-K-2021-AAAI,ri2021mluke}, and our work is another demonstration of their utility, particularly in sentence embeddings.
One promising future direction is exploring how to better exploit the cross-lingual nature of entities.

Our experiments also demonstrate the utility of Wikipedia as a multilingual database.
As described in Section~\ref{sec:case_study}, Wikipedia entity annotations can compensate for the lack of parallel resources in learning cross-lingual representations.
Wikipedia currently supports more than 300 languages, and around half of them have over 10,000 articles.\footnote{\url{https://meta.wikimedia.org/wiki/List_of_Wikipedias}}
Moreover, Wikipedia is ever growing; it is expected to include more and more languages.\footnote{\url{https://incubator.wikimedia.org/wiki/Incubator:Main_Page}}
This will motivate researchers to develop methods for multilingual models including low-resource languages in the aid of entity annotations in Wikipedia.

However, the reliance on Wikipedia for training data may limit the application of the models to specific domains (e.g., general or encyclopedia domains).
To apply EASE to other domains, one may need to annotate text from the domain either manually or automatically.
Future work can investigate the effectiveness of the entity CL in other domains and possibly its the combination with an entity linking system.

Finally, we note that the supervision signal in EASE is inherently noisy.
Different entities have different characteristics as a topic indicator, and sentences that contain the same entity do not necessarily share meaning.
Future work can address this by considering how an entity is used in a sentence to obtain more reliable supervision.

\section*{Acknowledgements}

This work was partially supported by JSPS KAKENHI Grants JP16H06302, JP18H04120, JP21H04907, and JP21K18023, and by JST CREST Grants JPMJCR18A6 and JPMJCR20D3, Japan.
JPMJCR18A6: JST CREST (FY2018-FY2023) PI Prof. Yamagishi
JPMJCR20D3: JST CREST (FY2020-FY2025) PI Prof. Echizen
JP16H06302: KAKENHI, Scientific Research (S) (FY2016-FY2020) PI Prof. Babaguchi, Co-PI Prof. Echizen
JP18H04120: KAKENHI, Scientific Research (A) (FY2018-FY2020) PI Prof. Echizen, Co-PI Prof. Yamagishi

<Follows are newly accepted.>
JP21H04907: KAKENHI, Scientific Research (A) (FY2021-FY2023) PI Prof. Echizen, Co-PI Prof. Yamagishi
JP21K18023: KAKENHI, Early-Career Scientists (FY2021-FY2023) PI Dr. Le Trung-Nghia

\bibliography{anthology}
\bibliographystyle{acl_natbib}

\newpage
\appendix
\onecolumn

\section{Training Details}
\label{appendix:training_details}

We implement our EASE model using \texttt{transformers}\footnote{https://huggingface.co/docs/transformers/index}
libraries.
For the monolingual settings, we use the STS-B development set as in \citep{gao-etal-2021-simcse}.
For multilingual settings, we use the STS-B and SICK-R development set. In this setting, we simply concatenate the entity-sentence paired data for all 18 languages and randomly sample from the concatenated data to construct batches.\footnote{In our preliminary experiments, we also tested a setting in which data in the same language were used within the same batch; we did not observe a consistent improvement in the performance of either the SimCSE or EASE models.}
In both settings, we train our model for one epoch, compute evaluation scores every 250 training steps on the development data, and keep the best model.
We conduct a grid-search for batch size $\in \{64,128,256,512\}$ and learning rate $\in \{3e-05, 5e-05\}$.
The chosen hyperparameters for each model is shown in Table~\ref{tab: ease_hyperparameters}.

\begin{table}[h]
	\begin{center}
	\scalebox{0.9}{ 
\begin{tabular}{l|ccccc}
\toprule
Model & Batch size & Learning Rate  \\ \midrule
SimCSE-mBERT\ba & 128 & 3e-05  \\
SimCSE-XLM-R\ba & 128 & 3e-05 \\
EASE-BERT\ba & 64 & 3e-05  \\
EASE-RoBERTa\ba & 128 & 5e-05 \\
EASE-mBERT\ba & 256 & 5e-05  \\
EASE-XLM-R\ba & 64 &  3e-05 \\
\bottomrule
\end{tabular}
}
\end{center}
\caption{Hyperparameters for experiment.}
\label{tab: ease_hyperparameters}
\end{table}

For the loss balancing term $\lambda$ and softmax temperature $\tau$ in the EASE models (\cref{seq: training_data}), we empirically find that $\lambda = 0.01$, $\tau = 100$ for the monolingual setting and $\tau = 10$ for the multilingual setting work well.

\paragraph{Computing Infrastructure}
We run the experiments on a server with AMD EPYC 7302 16-Core CPU and a NVIDIA A100-PCIE-40GB GPU.
The training of EASE takes approximately 1 hour.

\section{Pooling Methods for SimCSE and EASE}
\label{app: pooling_method}

We compare several pooling methods on both SimCSE and EASE in the multilingual setting:
[CLS] with MLP; [CLS] with MLP during training only; [CLS] without MLP; mean pooling.
Table~\ref{tab:pooler} shows the evaluation results based on the STS-B and SICK-R development set.

\begin{table}[h]
    \begin{center}
    \begin{tabular}{l|cc} \toprule
      Pooler & SimCSE & EASE  \\ \midrule
         {[}CLS{]} pooling & & \\
           \qquad  w/ MLP & 63.0 & 65.0 \\
           \qquad  w/ MLP (train) & 72.0 &73.3 \\
           \qquad w/o MLP & 72.0 & 73.4\\
           mean pooling & \textbf{72.1} & \textbf{73.8} \\ \bottomrule
    \end{tabular}
    \end{center}
    \caption{Average Spearman's correlation for different pooling methods for SimCSE and EASE in multilingual setting on STS-B and SICK-R development set.}
    \label{tab:pooler}
\end{table}

The mean pooling representation performs best on both models.
We thus use mean pooling on both models in Section~\ref{sec: multilingual_experiment}.

\newpage

\section{Parallel Sentence Mining}

We evaluate the multilingual sentence embeddings with the parallel sentence mining task using the BUCC 2018 shared task dataset \citep{zweigenbaum2018overview}.
The task is to find the parallel pairs given monolingual sentence pools in two languages, with 2–3\% of the sentences being parallel, to find the parallel pairs.

Each model uses the raw embedding output and performance is evaluated without fine-tuning.
We first encode all sentences into embeddings and compute the cosine similarity scores between all possible sentence pairs.
We then retrieve the sentence pairs with above a fixed threshold and compute the F1 score using the ground-truth parallel pairs.

As the test set is not publicly available, we use the sample set to tune the threshold of the parallel sentence mining and the training set for evaluation, which is a common practice in similar studies \citep{hu2020xtreme,Feng2020LanguageagnosticBS}.

The results are summarized in \Table{table:bucc}.
Our EASE models outperform the SimCSE baselines across the languages, demonstrating that the entity contrastive objective improves the alignment of the multilingual sentence embeddings without a parallel corpora.
However, performance is significantly poor than that of LaBSE, which is trained using massive amounts of parallel corpora, suggesting that we still need parallel resources to be competitive on this task.

\begin{table}[h]
\centering
\begin{tabular}{lcccc}\toprule
           & en-de & en-fr & en-ru & en-zh\\\midrule
SimCSE-mBERT\ba &  13.2 &  19.2 &  7.9  &  11.5\\
EASE-mBERT\ba &  26.9 &  33.8 &  24.2 &  32.9 \\ \midrule
SimCSE-XLM-R\ba &  31.8 &  32.3 &  28.9 &  19.9\\
EASE-XLM-R\ba &  33.3 &  33.2 &  33.6 &  23.4\\ \midrule
LaBSE &  89.0 &  88.2 &  84.7 &  74.2\\
\bottomrule
\end{tabular}
\caption{The F1 scores on BUCC 2018 the training set. Retrieval is performed in forward search, i.e., English sentences as the targets and the other language as the queries.}
\label{table:bucc}
\end{table}

\section{Detailed Settings for MLDoc Experiment}
\label{app: mldoc}

We use the english.train.1000 and english.dev datasets for the training and validation data, respectively. 
We conduct a grid-search for batch size $\in \{32,64,128\}$ and learning rate $\in \{0.1, 0.01, 0.001\}$ using validation data~\ref{tab: mldoc_hyperparameters}.
We run the experiment three times with different random seeds and record the average scores.

\begin{table}[h]
	\begin{center}
\begin{tabular}{l|ccccc}
\toprule
Model & Batch size & Learning Rate  \\ \midrule
mBERT\ba (avg.) & 32 & 0.1  \\
XLM-R\ba (avg.) & 32 & 0.1 \\
SimCSE-mBERT\ba & 32 & 0.1  \\
SimCSE-XLM-R\ba & 32 & 0.01 \\
EASE-mBERT\ba & 32 & 0.01  \\
EASE-XLM-R\ba & 32 &  0.01 \\
\bottomrule
\end{tabular}
\end{center}
\caption{Hyperparameters for MLDoc experiment}
\label{tab: mldoc_hyperparameters}
\end{table}

\newpage

\section{Construction of MewsC-16 Dataset}
\label{app: MewsC-16}

To construct the MewsC-16 dataset, we collect sentences for each category in each language from the Wikinews dump.\footnote{\url{https://dumps.wikimedia.org/backup-index.html}}
We first select 13 topic categories in the English Wikinews~\footnote{\url{https://en.wikinews.org/wiki/Category:News_articles_by_section}} that are also defined in other languages (Science and technology, Politics and conflicts, Environment, Sports, Health, Crime and law, Obituaries, Disasters and accidents, Culture and entertainment, Economy and business, Weather, Education, Media).
We then collect pages with topic categories for each language and remove the pages with two or more topic categories.
We clean the text on each page with the \texttt{Wikiextractor} tool\footnote{\url{https://github.com/attardi/wikiextractor}}, and split it into sentences using the \texttt{polyglot} sentence tokenizer.
Finally, we use the first sentence assuming that it well represents the topic of the entire article~\cite{10.1147/rd.24.0354,10.1145/321510.321519}.
The corpus statistics for each language are shown in Table~\ref{tab: statistics_for_MewsC-16}.

\begin{table}[h]
	\begin{center}

\begin{tabular}{crr|crr}
\toprule
Language & \# of sentences  & \# of label types & Language & \# of sentences  & \# of label types  \\
\midrule
ar & 2,243 & 11 & fr & 10,697 & 13 \\
ca & 3,310 & 11 & ja & 1,984 & 12 \\
cs & 1,534 & 9 & ko & 344 & 10 \\
de & 6,398 & 8 & pl & 7,247 & 11 \\
en & 12,892 & 13 & pt & 8,921 & 11 \\
eo & 227 & 8 & ru & 1,406 & 12 \\
es & 6,415 & 11 & sv & 584 & 7 \\
fa & 773 & 9 & tr & 459 & 7 \\
\midrule
   &     &   & total & 65,425 & 13 \\
\bottomrule
\end{tabular}
\end{center}
\caption{Corpus statistics for MewsC-16}
\label{tab: statistics_for_MewsC-16}
\end{table}

\newpage
\section{Baselines}

For average GloVe embedding~\cite{pennington-etal-2014-glove}, we use open-source GloVe vectors trained on Wikipedia and Gigaword
with 300 dimensions.\footnote{\url{https://nlp.stanford.edu/projects/glove/}}
We use the pretrained model from HuggingFace’s \texttt{Transformers}\footnote{\url{https://github.com/huggingface/transformers}} for vanilla pretrained language models, including BERT (bert-base-uncased)~\cite{devlin-etal-2019-bert}, RoBERTa (roberta-base)~\cite{liu2019roberta}, mBERT (bert-base-multilingual-cased) and XLM-R (xlm-roberta-base)~\cite{conneau-etal-2020-unsupervised}.
We use the published checkpoints for unsupervised SimCSE~\cite{gao-etal-2021-simcse}\footnote{\url{https://github.com/princeton-nlp/SimCSE}}, CT~\cite{DBLP:conf/iclr/CarlssonGGHS21}\footnote{\url{https://github.com/FreddeFrallan/Contrastive-Tension}}, and DeCLUTR~\cite{giorgi-etal-2021-declutr}.\footnote{\url{https://github.com/JohnGiorgi/DeCLUTR}}

\section{Monolingual STS and STC}
\label{app: monolingul_full_results}

Table \ref{tab: mono-sts-full} and \ref{tab: mono-stc-full} show the complete results for seven STS tasks and eight STC tasks.
For STS, the average EASE performance is slightly better than that of SimCSE, although the advantage is not consistent across tasks.
For most of the STC tasks, EASE consistently outperforms SimCSE.
These results indicate that EASE stands out at capturing high-level categorical semantic structures and that its ability to measure sentence semantic similarity is comparable to or better than that of SimCSE.

\begin{table*}[h]
    \begin{center}
    \scalebox{0.9}{ 
    \begin{tabular}{l|cccccccc}
       \toprule
       {Model} & {STS12} & {STS13} & {STS14} & {STS15} & {STS16} & {STS-B} & {SICK-R} & {Avg.} \\ \midrule
        GloVe embeddings (avg.) & 55.1 & 70.7 & 59.7 & 68.3 & 63.7 & 58.0 & 53.8 & 61.3 \\
        BERT\ba~(avg.) & 30.9&59.9&47.7&60.3&63.7&47.3&58.2&52.6 \\
        BERT\ba-flow & 58.4&	67.1&	60.9&	75.2&	71.2&	68.7&	64.5&	66.6 \\ 
        BERT\ba-whitening & 57.8& 66.9 & 60.9 & 75.1& 71.3& 68.2& 63.7& 66.3\\ 
        IS-BERT\ba$^\heartsuit$ & 56.8 & 69.2 & 61.2 & 75.2 & 70.2 & 69.2 & 64.3 & 66.6 \\
        CT-BERT\ba & 61.6 & 76.8 & 68.5 & 77.5 & 76.5 & 74.3 & 69.2 &72.1 \\ 
        SimCSE-BERT\ba & 68.4&	\textbf{82.4} &	\textbf{74.4}&	80.9&	78.6&76.9& \textbf{72.2}&	76.3 \\
        \textbf{EASE-BERT\ba} &\textbf{72.8}&81.8&73.7&\textbf{82.3}&\textbf{79.5}&\textbf{78.9}&69.7&\textbf{77.0}\\ 
        \hline
        RoBERTa\ba~(avg.) & 32.1&56.3&45.2&61.3&62.0&55.4&62.0&53.5 \\ 
        RoBERTa\ba~(first-last avg.) & 40.9&58.7&49.1&65.6&61.5&58.6&61.6&56.6  \\ 
        DeCLUTR-RoBERTa\ba & 52.4 & 75.2& 65.5 & 77.1 & 78.6 & 72.4 & \textbf{68.6}& 70.0\\
        SimCSE-RoBERTa\ba  & 68.7&	\textbf{82.6}&	\textbf{73.6}&81.5&	\textbf{80.8}&	\textbf{80.5}&	67.9&	76.5 \\
        \textbf{EASE-RoBERTa\ba} & \textbf{70.9}&81.5&73.5&\textbf{82.6}&80.5&80.0&68.4& \textbf{76.8} \\ \bottomrule
     \end{tabular}
     }
     \end{center}
     	\caption{Spearman’s correlation for monolingual semantic textual similarity tasks.}
     \label{tab: mono-sts-full}
\end{table*}

\begin{table*}[h]
    \begin{center}
    \scalebox{0.9}{ 
    \begin{tabular}{l|ccccccccc}
       \toprule
       {Model} & {AG}  & {Bio} &  {G-S} & {G-T} & {G-TS} & {SO} & {SS}  & {Tweet} & {Avg.}\\ \midrule
        GloVe embeddings (avg.) & 83.2&30.7&59.0&58.3&67.4&29.9&70.4&52.1&56.4 \\
        BERT\ba~(avg.) &79.8&32.5&55.0&47.0&62.4&21.7&64.0&44.6&50.9 \\
        CT-BERT\ba & 79.2&\textbf{38.7}&\textbf{65.5}&\textbf{60.7}&\textbf{69.8}&67.9&55.5&\textbf{55.2}&61.6 \\ 
        SimCSE-BERT\ba &74.4&34.3&59.5&57.8&64.4&49.6&64.3&52.1&57.1 \\
        \textbf{EASE-BERT\ba} & \textbf{85.8}&36.2&60.5&60.4&67.0&\textbf{68.1}&\textbf{71.7}&54.8&\textbf{63.1} \\ 
        \hline
        RoBERTa\ba~(avg.) & 66.5&26.6&47.9&42.8&58.3&16.7&30.0&38.6&40.9 \\ 
        DeCLUTR-RoBERTa\ba & \textbf{80.7}&\textbf{41.0}&\textbf{65.2}&\textbf{60.5}&\textbf{69.6}&32.9&\textbf{73.6}&\textbf{56.8}&\textbf{60.0}\\
        SimCSE-RoBERTa\ba & 69.8&37.3&60.0&58.0&66.6&69.3&48.3&50.0&57.4\\
        \textbf{EASE-RoBERTa\ba} & 69.4&39.3&60.7&57.7&66.3&\textbf{73.9}&49.4&51.8&58.6\\ \bottomrule
     \end{tabular}
     }
     \end{center}
     \caption{Clustering accuracy for monolingual short text clustering tasks.}
     \label{tab: mono-stc-full}
\end{table*}




\end{document}